\documentclass[letterpaper, 10 pt, conference]{ieeeconf}
\IEEEoverridecommandlockouts
\usepackage{cite}
\usepackage{amsmath,amssymb,amsfonts}
\usepackage{algorithmic}
\usepackage{textcomp}
\usepackage{xcolor}
\usepackage{wrapfig}
\usepackage{multicol}
\usepackage{booktabs}
\usepackage{hyperref}
\usepackage{float}
\usepackage{stfloats}
\usepackage{caption}
\usepackage{graphicx}

\def\BibTeX{{\rm B\kern-.05em{\sc i\kern-.025em b}\kern-.08em
    T\kern-.1667em\lower.7ex\hbox{E}\kern-.125emX}}
\begin{document}

\title{QTOS: An Open-Source Quadruped Trajectory Optimization Stack}

\author{Alexy Skoutnev$^{1}$, Andrew Cinar$^{1}$, Praful Sigdel$^{1}$, and Forrest Laine$^{1}$
\thanks{$^{1}$Vanderbilt University}%
}

\maketitle

\begin{abstract}
We introduce a new open-source framework, Quadruped Trajectory Optimization Stack (QTOS), which integrates a global planner, local planner, simulator, controller, and robot interface into a single package. QTOS serves as a full-stack interface, simplifying continuous motion planning on an open-source quadruped platform by bridging the gap between middleware and gait planning. It empowers users to effortlessly translate high-level navigation objectives into low-level robot commands. Furthermore, QTOS enhances the stability and adaptability of long-distance gait planning across challenging terrain. Additional videos and materials can be found at \href{https://alexyskoutnev.github.io/Quadruped-Trajectory-Optimization-Stack}{https://alexyskoutnev.github.io/Quadruped-Trajectory-Optimization-Stack}.
\end{abstract}

\section{Introduction}

In this paper, we introduce the Quadruped Trajectory Optimization Stack (QTOS), which is a software stack that spans motion planning, control software, hardware interfaces, and deployment tools. QTOS is a one-stop framework for simplifying the setup and use of open-source quadruped robots for research. Many high-performance quadrupeds such as Unitree A1 \cite{semini2015versatile}, Anymal \cite{hutter2016anymal}, and Spot \cite{BostonDynamics_2023}, can be expensive for a new research lab, and the attached proprietary software can make comprehensive system testing difficult \cite{boney2022realant}. Because of this, open-source quadrupeds are gaining popularity, yet a critical gap exists in readily available open-source software for designing, verifying, and implementing advanced legged locomotion \cite{techxplore-legged-robot-2022}. 

Many popular open-source projects have tried providing an end-to-end framework for quadruped control, and navigation such as Quad-SDK \cite{norby2022quad-sdk}, WoLF \cite{raiola2022wolf}, but none have been built down to directly communicate with an open-source quadruped system. Open-source quadruped robots, such as, SOLO \cite{grimminger2020a}, Oncilla \cite{sproewitz2018oncilla}, and Stanford doggo \cite{kau2019stanforddoggo}, facilitate collaboration and expedite troubleshooting for quadruped hardware issues, and lead to accelerated development cycles and robust hardware frameworks through community-driven effort \cite{open-source}.

To address this need, we present the Open-source Quadruped Trajectory Optimization Stack (QTOS) framework, emphasizing its role as a comprehensive solution for simplifying advanced motion planning on the quadruped platform, SOLO12. QTOS provides intuitive tools for translating high-level navigation tasks into low-level commands, with a design emphasis on simplicity and portability. With the development of QTOS, we built a modern locomotion framework by integrating a state-of-the-art gait planner with a newly constructed global motion planner and end-effector controller that reliably generates and tracks long-distance stitched gait plans. Along the way, we addressed the community's need for a standardized workflow that spans planning, control software, hardware interface, and deployment tools \cite{cheetharobot}.

\begin{figure}[t!]
  \centering
  \includegraphics[width=\linewidth, height=5cm]{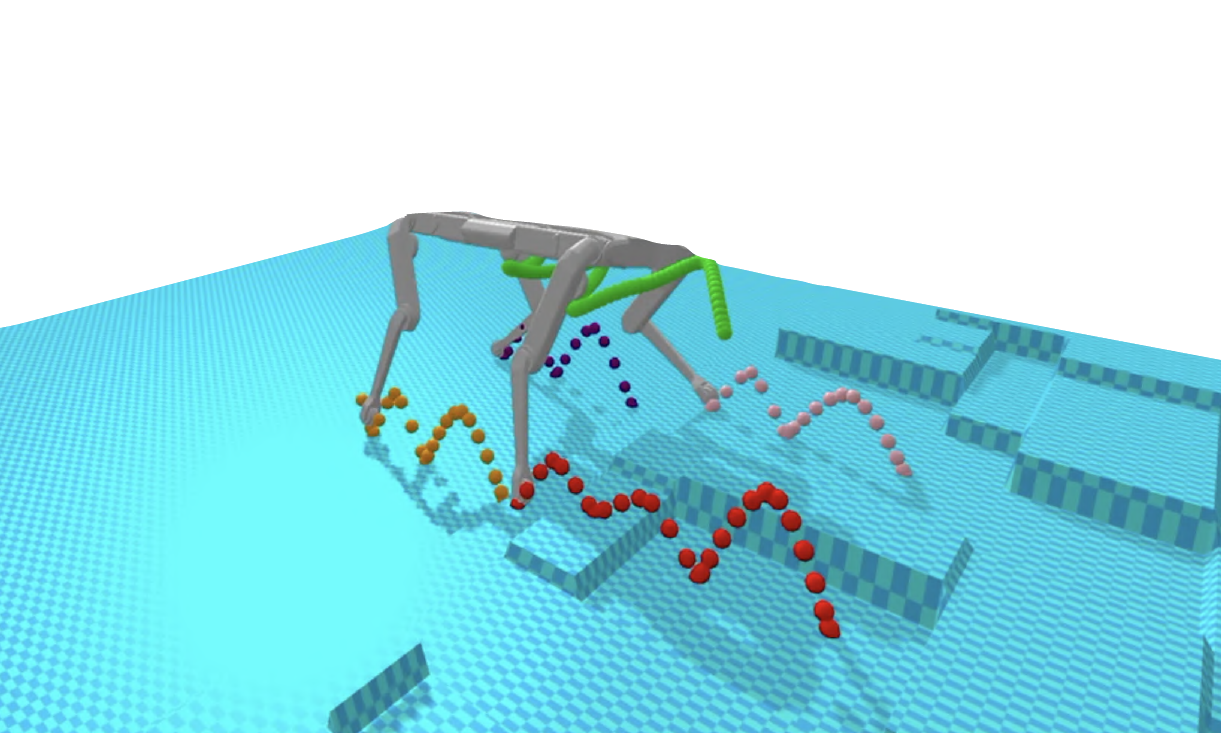}
  \caption{Trajectory generated by the QTOS local planner while performing a climbing task. In this environment, the robot successfully overcomes different challenging terrains in simulation, showcasing the adaptability of QTOS.}
\end{figure}
QTOS was developed to address the practical demand for a workflow that facilitates the design and the implementation of complex legged locomotion planning tasks following the setup of open-source quadruped hardware. To the best of our knowledge, SOLO is the only open-source robot that is able to execute dynamic motion plans and was used as the testing platform featured in this paper \cite{grimminger2020a}. Specifically, SOLO12 by PAL Robotics is an open-source torque-controlled quadruped robot system that uses off-the-shelf components such as high-torque brushless DC motors and 3D printed components, developed by The Open Dynamic Robot Initiative (ODRI) \cite{grimminger2020a}. SOLO12 unifies open-source hardware, firmware, and middleware within a single ecosystem. QTOS plays a crucial role in bridging the gap between the middleware and the realization of dynamic motion planning, effectively addressing the practical requirements that arise along the way by offering an integrated platform.

Our work has three main contributions: (1) We introduce QTOS as a full-stack interface for planning and control on quadruped systems. (2) We have developed a software toolkit and a streamlined workflow aimed at expediting the testing and development cycle for the SOLO12 robot. Finally, (3) we demonstrate the effectiveness of stitched motion planning facilitated by QTOS in various simulation tasks and discuss the reliability of long-distance stitched gait plans.

\section{Background and Overview}

There are significant gaps in existing features in the quadruped ecosystem to get from working hardware to implementing advanced motion plans on an open-source quadruped robot. We have extended and improved features from a range of open-source code such as ODRI \cite{grimminger2020a}, towr \cite{winkler18}, and RViz \cite{Rviz}, and we introduced new features aimed at simplifying the physical realization of legged locomotion tasks.

\subsubsection{Cross-Platform Support}

Many robotics-based packages depend on Linux-based headers, which can inadvertently exclude potential users. To facilitate cross-platform accessibility, we employed both the containerization tool Docker \cite{erkel2014docker} and the package management system Anaconda \cite{anaconda}.

\subsubsection{UX Extension and Physical Hardware Support}

After assembling the robot, calibrating the incremental encoders in each limb is crucial. This is achieved by sweeping and saving the index pulse positions. We noticed that the calibration procedure provided in the ODRI software is time-consuming, and often results in a misaligned starting state. QTOS streamlines the necessary setup phase for robot operation by employing a state machine and an improved sweeping algorithm, resulting in an automated and convenient calibration procedure. In the common case of detecting the wrong index pulse, as we will discuss later, we feature command line arguments for a straightforward compensation for the misaligned joints. Furthermore, we needed increased diagnostics monitoring features required to track the robot's health, and we have introduced improved real-time data monitoring for user experience (UX). The improved UX enables users to assess controller performance and vital robot state statistics. This includes information on packet communication and controller operational states. Moreover, we have enhanced the user interface to improve the overall user experience for testing and evaluating trajectories and efficiently utilizing the software's capabilities. These enhancements streamline the process of initiating and executing trajectories through the implementation of intuitive state-based user commands.

\subsubsection{Unified, Improved, Refactored SDK}
The original SOLO12 SDK relies on multiple external dependencies such as ROS2 \cite{doi:10.1126/scirobotics.abm6074} and many sparsely connected repositories in the ODRI ecosystem. While these dependencies can extend the functionality of the software, they also introduce complexity and potential compatibility issues \cite{google}, \cite{cox2019software}. This complexity can make it challenging for users to set up and maintain their robotic systems efficiently. To address this, we have improved the original source code by eliminating dependencies and functions and refining vital methods. As a result, we have consolidated the SDK functionality into a single executable. 

All of the extended and improved SDK and extended functionality have been consolidated into the \href{https://github.com/Alexyskoutnev/SOLO12_SDK}{SOLO12{\textunderscore}SDK} repository. \href{https://github.com/Alexyskoutnev/SOLO12_SDK}{SOLO12{\textunderscore}SDK} provides a comprehensive and cohesive toolkit for users working with the SOLO12 robot.

\subsection{Stitched Motion Planning}

An integral module of the QTOS is the continuous trajectory optimizer used to generate gait plans for the SOLO12 robot. We discovered TOWR \cite{winkler18} as the preferred performance-efficient gait planning solver without compromising reliability as seen by other solvers that decompose the optimization problem into several sub-problems \cite{bjelonic2021wholebody}. TOWR is able to automatically determine the gait sequence, step-timing, footholds, end-effector motions, and 6-DoF body coordinates without the need to solve a mixed integer or a complementary problem \cite{Zelinsky} \cite{corrigendum2014direct}. Its flexibility in regulating the gait timing sequences is vital for constructing stable foot placements on elevated surfaces and allows for complete high-level autonomy. However, TOWR is prone to producing infeasible gaits for long-distance navigation tasks and requires intricate motion composition to robustly produce online trajectories. In this work, we discover that online trajectory planning is possible without the need for a complex MPC controller \cite{bjelonic2021wholebody}. Having a dynamic look-ahead process that targets stable state configurations, we utilize trajectory stitching to produce robust gaits for real-time motion planning.

\section{Software Architecture}

\begin{figure}[htp!]
  \includegraphics[width=\linewidth, height=8cm]{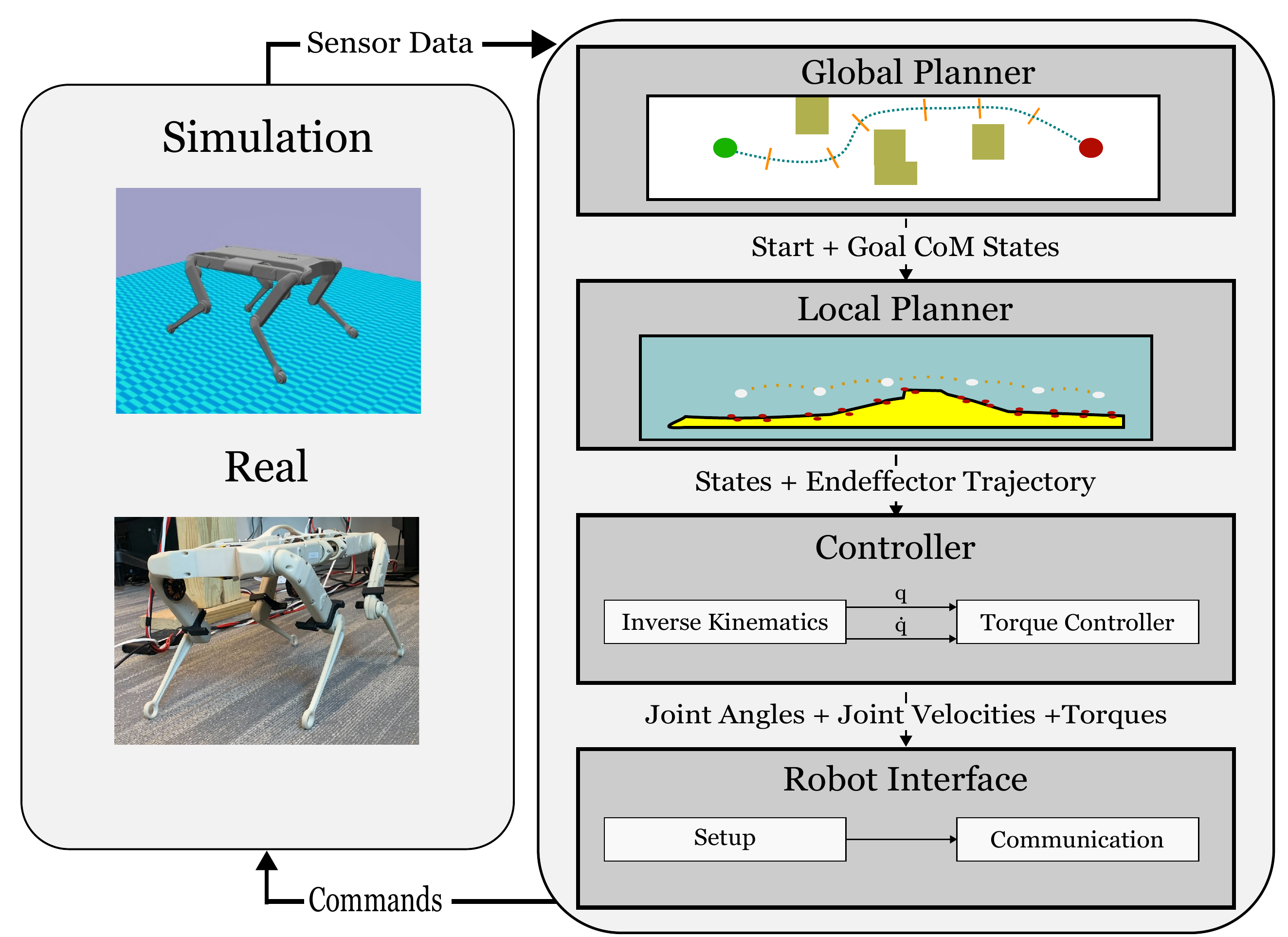}
      \caption{The QTOS system architecture follows a hierarchical structure where a high-level navigation task is translated into low-level robot commands.}
      \label{fig:QTOS}
\end{figure}

We introduce a new open-source framework coined, QTOS, that integrates a global planner, local planner, simulator, controller, and robot interface all in one package. To complete the end-to-end integration, QTOS also includes tools for hardware calibration, diagnostic monitoring, and an intuitive user interface to operate the robot. The full-stack software framework streamlines quadruped planning, control, simulation, and communication for researchers and engineers, allowing them to emphasize core algorithm development rather than software tooling and infrastructure. Consequently, QTOS follows a modular structure that facilitates customization, as shown in Figure \ref{fig:QTOS}. A high-level navigation task is parsed through four layers of the stack before being consumed by the robot as a low-level data package.

\subsection{Global Planner}

\begin{figure}[htp!]
  \includegraphics[width=\linewidth,height=3cm]{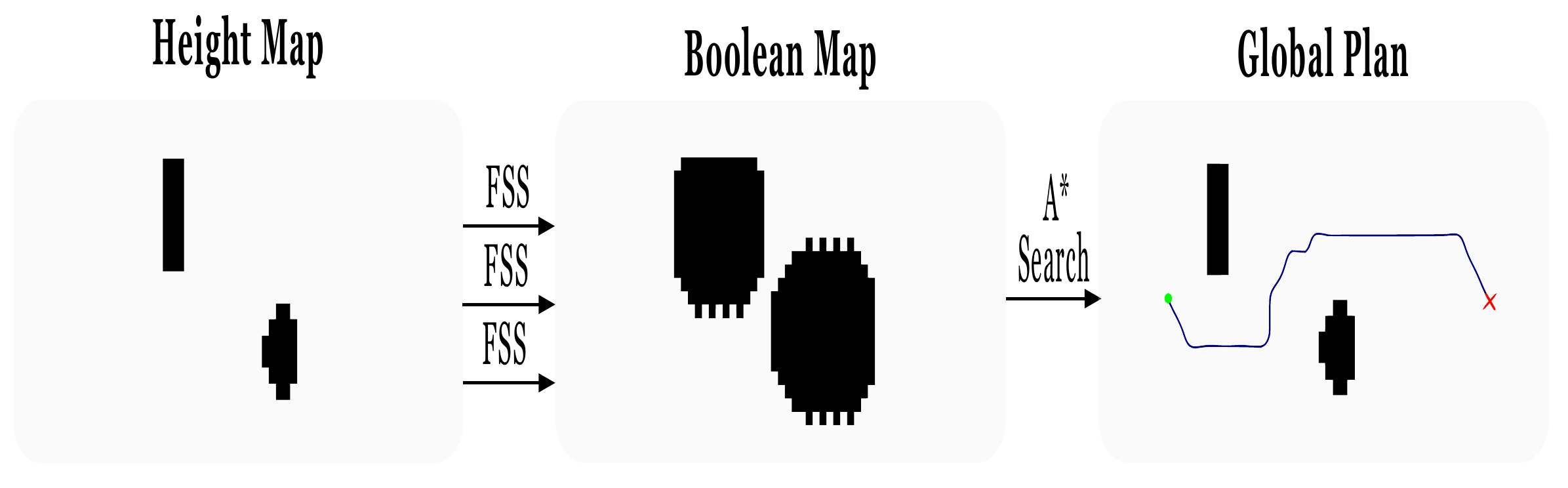}
    \centering
      \caption{The global trajectory plan is computed via a 2-step transformation procedure. The FSS algorithm concurrently processes the entire environment until it is transformed into a Boolean map. In the last step, a tailored A* search is performed and a global trajectory spline is constructed. The three maps were generated from QTOS. }
      \label{fig:fss}
\end{figure}

At the top of the stack is the global planner, responsible for computing a global trajectory that can be divided into solvable segments for the local planner. The global planner guides the robot from the initial state to a user-provided goal state while promoting feasibility for each local trajectory plan through a dynamic look-ahead search process. At each update step, this search process identifies a future trajectory state configuration where all end-effectors are in contact with the ground, serving as the new starting configuration state. Terrain information is supplied to the global planner via a 2.5D height map, corresponding to the robot's working environment. As part of its processing, the global planner simultaneously searches for and determines a feasible and nearly optimal trajectory plan. Regions potentially deemed infeasible, such as walls or steep curves, are evaluated using a custom feasibility search strategy (FSS), illustrated in Figure \ref{fig:fss}.

The FSS algorithm performs a grid search for groups of nodes that violate the end-effector kinematics height constraint. The violation is determined by the height difference between a test node and its neighbors and if it is greater than a platform-determined height-deviation parameter. Then, FSS creates a convex hull that encloses the group of potentially infeasible nodes and utilizes multi-threading to pair bilateral starting and goal points inside or near the convex hull. Each pair is tested as a micro-trajectory (trajectories with a magnitude less than 0.2 meters) and assigned true or false based on the feasibility state of the local planner. In this manner, the original 2.5D height map is transformed into a Boolean grid map which is then made available to a tailored A* search algorithm that finds the shortest path and converts it to a cubic spline. Keep in mind a different type of feasibility test could be performed in the FSS algorithm based on the user's preferences for speed and accuracy.

The global trajectory plan is divided into multiple local gait plans, and this division is governed by a step-size parameter that controls the lookahead distance along the global trajectory spline. During runtime, the current local gait plan is actively tracked, while in the background, the next local gait plan and the concluding segment of the previous trajectory plan undergo optimization by the trajectory solver. Once the next plan is completely solved, the global planner stitches together these two plans, considering the robot's current state. This process repeats with the new next plan. The global planner operates at the same rate as a full gait cycle (0.5 Hz) and can react to disturbances or unexpected obstacles by adjusting the start-goal state pair for the subsequent gait plan.

\subsection{Local Planner}

The local planner generates a gait trajectory based on the initial and destination states in $\mathrm{SE}(3)$, which are provided by the global planner. At each node along the trajectory, the end-effector and 6-DoF body motion are calculated using a modified TOWR solver. The modified TOWR executable can accommodate custom height map constraints specified by the user-defined terrains through the QTOS map generation API. Additionally, it was discovered that using a custom gait pattern significantly enhances the stability of the gait trajectory for the lower-level modules of the stack, contributing to the generation of more feasible trajectories. To ensure broad accessibility, QTOS encapsulates the Linux executable of TOWR within a Docker container \cite{erkel2014docker}, enabling platform-independent API calls between the user's system and the trajectory solver. The TOWR container achieves an update frequency of 2 Hz.

\subsection{Controller}

The controller computes the desired joint angle and joint velocity, or the desired torque command for the embedded controller located inside the robot. Each trajectory point is processed through an inverse kinematics solver and transformed into a torque command by a PD controller. The controller has an update frequency of 1000 Hz.

\subsubsection{Inverse Kinematics}

We use the Damped Least Squares (DLS) method \cite{damped} to find the generalized reference joint space position $q_\text{ref}$ and reference joint space velocity $\dot{q}_\text{ref}$.
At each time step $t_n$, the difference between the desired leg state $x_\text{des} \in R^3$, and the current leg state $x_\text{cur} \in R^3$ is calculated as the error vector $e = x_\text{des} - x_\text{cur}$. Likewise, the end-effector Jacobian $J_e$ for the legs is computed and is used to obtain the least squared error solution. 
\begin{align}
    \dot{q}_\text{ref} = J_e^{\dagger} e 
\end{align}
Here, $J_e^{\dagger}$ refers to the pseudo-inverse matrix. Afterward, the joint space position is updated with the damping factor $\lambda$,
\begin{align}
    q_\text{ref} = q_\text{ref} + \lambda \dot{q}_\text{ref}
\end{align}
The DLS method is repeated until error $e$ is sufficiently small or the maximum number of interactions is reached. Keep in mind that joint-velocity control was used over joint-position control in the control software because of two reasons. Firstly, joint-velocity control enhances dynamic and responsive feedback \cite{kelly2005manipulator}. Secondly, and notably, joint-velocity control effectively mitigates drift in long-distance gait plans.

\subsubsection{PD Controller}

The PD controller provides the feedback torque needed to realize the desired trajectory by the following control law given the system states $q$ and $\dot{q}$,
\begin{align}
    \tau_\text{ref} = K_p (q_\text{ref} - q) + K_d (\dot{q}_\text{ref}- \dot{q})
\end{align}
where $K_p$ and $K_d$ are the proportional and derivative gains respectively.

\subsection{Robot Interface}
The robot interface is divided into multiple components: (1) The setup component which is responsible for configuring the physical robot to be in a ready-to-run state, (2) the communication component which handles the data packages between the user's system and the ESP32 microprocessor masterboard \cite{odi_masterboard}, (3) system diagnostics and data collection tools for debugging trajectory plans and other hardware-related issues.

\subsubsection{Setup Component}

\begin{figure}[t!]
    \includegraphics[width=.8\linewidth, height=4cm]{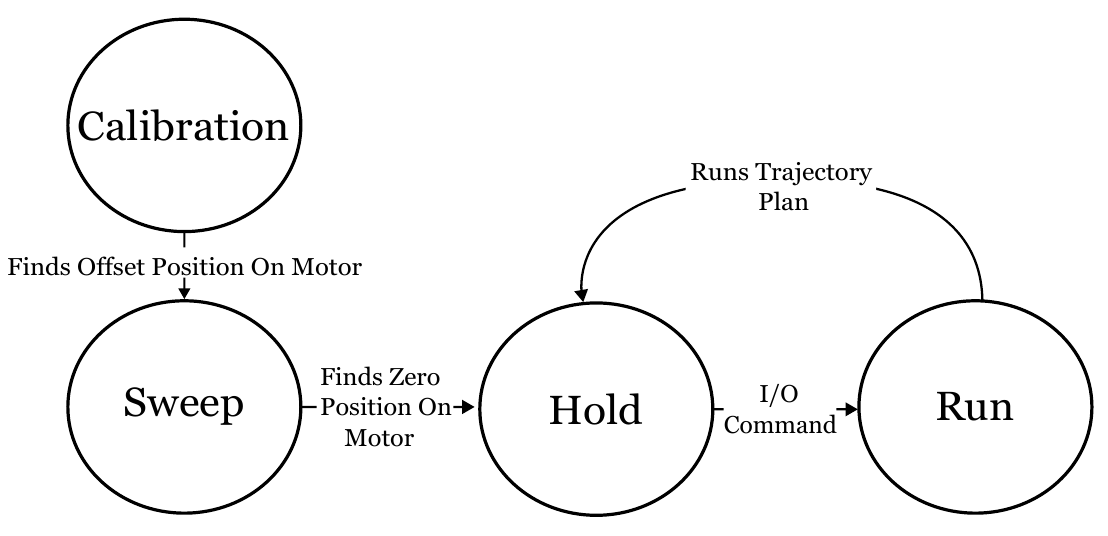}
     \centering
      \caption{The state machine described above configures the quadruped robot into the Run state. Calibration via Sweep to find the desired zero positions of all joints. After the Sweep is complete and upon user command, it transitions to Hold, tracking the initial starting state. Finally, upon user command, it enters the Run state to execute the trajectory plan.}
      \label{fig:state_machine}
\end{figure}

Before performing any tasks, the SOLO12 requires an initial calibration process because it is not an absolute encoder. The calibration procedure requires the determination of the nearest encoder index pulse position with respect to a desired reference zero position, we refer to this as the ``hard" calibration procedure. Hard calibration needs to happen once after the hardware has been put together, or after there have been changes to the hardware. After the hard calibration has been completed once, we need to determine the nearest index pulse with respect to the initial pose of the robot every time the quadruped robot is powered up. We do this by sweeping the joints near their initial positions (the pose of the robot when the power is turned on), and we refer to this process as the ``soft" calibration procedure. SOLO12 actuators have a reduction ratio of 9:1, which means one full rotation of the motor shaft is equivalent to 9 index pulses on the encoder. We approximately bring the joints to their desired reference positions to consistently calibrate the quadruped robot to the reference position. The sweeping algorithm is based on a sinusoidal function, $f(q, t)$, that oscillates between minimum and maximum joint space values $q \in [0, 2\pi/9]$ for each robot actuator. The current joint angle $q$, max-angle search amplitude $A$, run time $t$, and the sign function, 
$$\mathrm{sign(x)} := 
\begin{cases}
    x  & \text{Joint axis is clockwise} \\
    -x & \text{Joint axis is counter-clockwise} 
\end{cases}
$$ 
is used to determine the next reference joint position $q_\text{ref}$.  
\begin{equation}
q_\text{ref} = f(q, t) = 9 q [A - A \cos(2\pi t)] \,\mathrm{sign}(q)
\end{equation}
Once the encoding zeros are found and recorded, the robot is configured to hold state where the robot tracks the initial state of the trajectory plan. Upon receiving the user's I/O command, it transitions to the run state where the trajectory plan is executed. The setup sequence is represented by a finite state machine as shown in Figure \ref{fig:state_machine}. Note that the I/O command can also be executed automatically for online trajectory generation. 

Even after a successful hard calibration, during the subsequent soft calibration attempts, we observe that some joints may be offset by the distance between two index pulses due to the reduction ratio. For a successful soft calibration, the joints initially must be put within $\pm\pi/9$ radians ($\approx \pm 20^\circ$) of the desired reference position. Due to gravity and other factors, it is common for one or two encoders to fall outside of this range at power-on, which results in misaligned starting positions. To correct this misalignment, we manually specify the desired index pulse offsets in the command line arguments provided to the robot interface.

\subsubsection{Communication Component}

The communication component is responsible for sending and receiving packets between the computer and the ESP32 microprocessor masterboard. The SOLO12 masterboard has two modes of
control: (1) On-board PD control, and (2) torque control. The communication component communicates the desired positions and velocities in the case of the on-board PD control, and the desired torques in the case of torque control. The on-board PD controller tracks the desired joint positions and velocities, or torques, using user-specified gains. This communication occurs at 1000 Hz and it is important to be carried out in a real-time manner for the accurate realization of the desired trajectory. However, the user interface runs on a user operating system, and the real-time behavior of the interface and the communication component needs to be monitored to ensure timing accuracy. To this end, the user interface communication component keeps track of timing statistics and prints them on screen for the user to verify that the communication component is running practically in real-time. In this case, ``practically" real-time means that every update from receiving sensor data, processing, and the sent commands are completed within the 1 ms time-frame. 

\subsection{User Tools}

QTOS contains user tools to facilitate rapid development and hardware testing in both simulation and real hardware. We present these tools in this subsection.

\subsubsection{Simulator}
\begin{figure*}[!t]
  \centering
  \includegraphics[width=\linewidth, height=6.5cm]{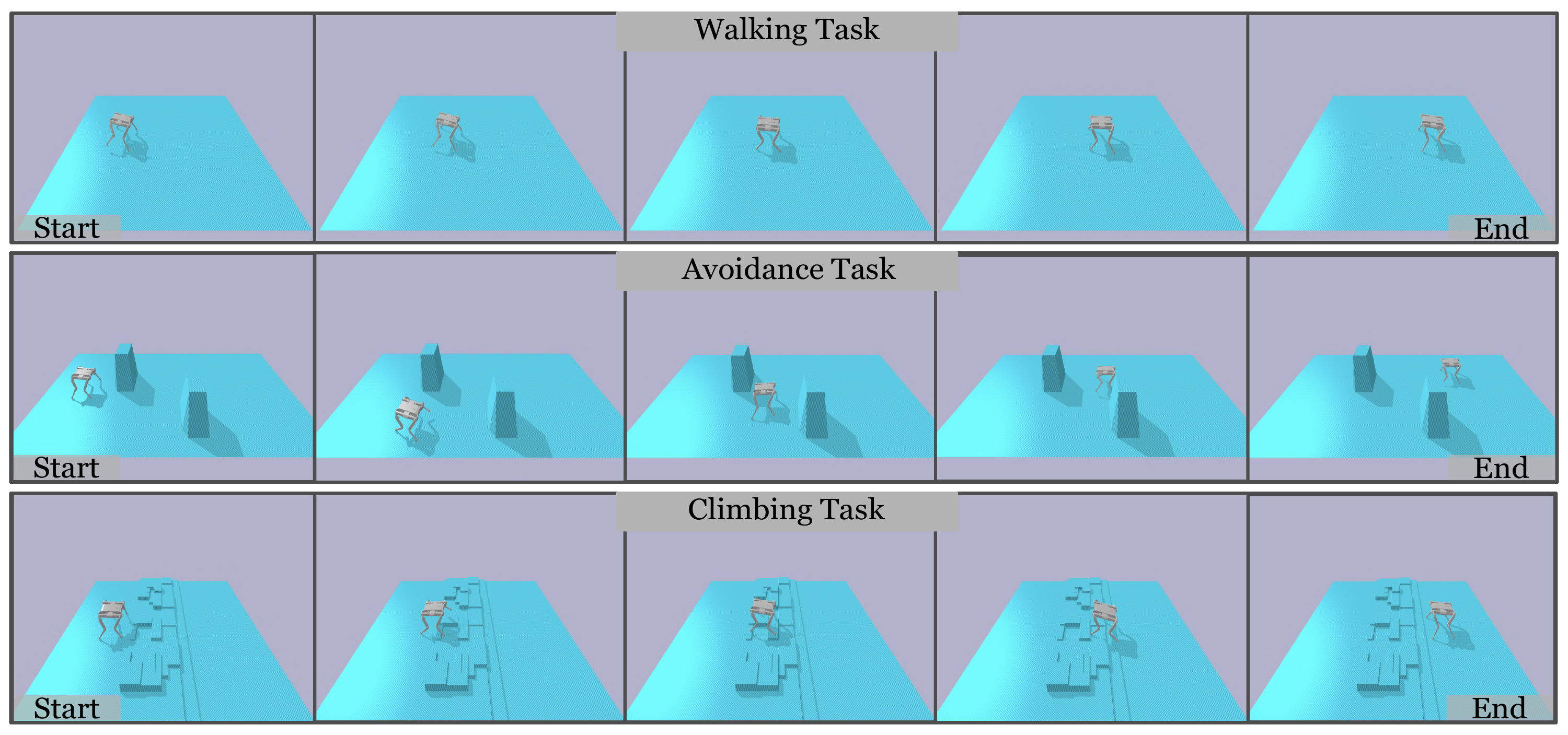}
  \caption{Timelapse of deploying QTOS in simulation. The three tasks: (1) walking, (2) avoidance, and (3) climbing can be seen from top to bottom respectively.}
  \label{fig:terrain_map}
\end{figure*}

The original TOWR trajectories were tested with RViz, but the absence of a physics engine made it challenging to assess the feasibility of the generated gait plans. During the design of QTOS, we prioritized a convenient and accurate simulation tool, leading to the integration of a physics simulator, Pybullet \cite{coumans2019}. With QTOS's simulator, users have the capability to create terrain environments tailored to their workspace, complete with visual modeling and trajectory tracking. The custom terrain environment is represented as an $n$ by $n$ grid of heights. Figure \ref{fig:terrain_map} illustrates an example terrain map file that users can edit, and it displays the resulting terrain.

\begin{figure}[t!]
  \includegraphics[width=\linewidth]{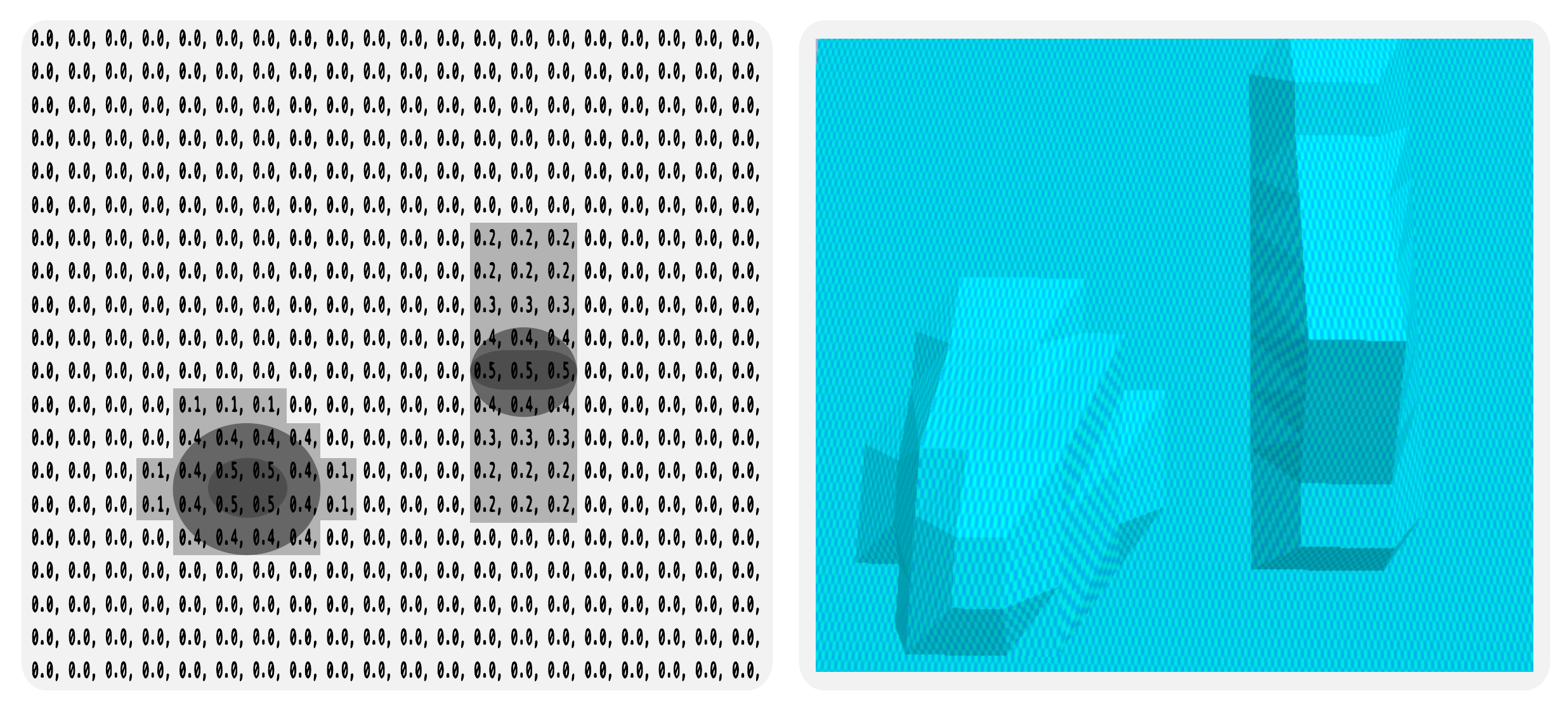}
      \caption{An example height map file being converted to a virtual terrain profile in the simulator. The user-defined inputs are highlighted with grey shading.}
\end{figure}

\subsubsection{Data Monitoring Tools} 

QTOS offers data monitoring tools for real-time analysis of critical performance metrics. The global plan is visualized as a Matplotlib plot \cite{matplotlib}. Additionally, reference states are logged alongside operational states, and these data sets are graphically represented using Matplotlib plots to evaluate the controller's tracking performance.

\section{Experiments}
In this section, we evaluate the performance of QTOS gait trajectories in the context of three key navigation tasks: (1) walking, (2) climbing, and (3) collision avoidance. We primarily focus on the performance and reliability of the stitched motion plans. Furthermore, we describe how QTOS handles stitched trajectories in order to determine the maximum time that QTOS can reasonably track. We aim to address the following questions using the experimental data: (1) How well does QTOS perform across a diverse set of navigation tasks using stitched motion plans? (2) What kind of behavior can we expect from QTOS when generating long-distance gait plans?
 
\subsection{Navigation Experiments}

In our navigation experiments, we define success as the robot's ability to navigate from the starting point to the specified goal without falling over. Failure occurs if the robot topples over or if it leaves the map boundary. To assess the consistency of the plans, we calculate the average distance traveled for each task. Additionally, we conduct a performance analysis by computing the average tracking error per second between the reference center of mass $x_\text{ref}$ and the realized center of mass $x_\text{real}$ using $n$ data points at a tracking rate $f_\text{track}$ of 1000 Hz.
$$\text{Tracking Error Rate} := \frac{\|x_\text{ref} - x_\text{real}\|_2}{n} * f_\text{track}$$
With the built-in Pybullet simulator, QTOS is benchmarked over 20 simulated runs where each environment is randomly generated with the following navigation tasks except the walking task.

\begin{enumerate}
    \item \textbf{Walking:} The robot is tasked with walking in a straight line, covering a distance of 2 meters while maintaining stability and balance.
    \item \textbf{Avoidance:} The robot successfully navigates around two obstructive walls while covering a distance of 2 meters.
    \item \textbf{Climbing:} The robot climbs over elevated terrain in its path while ensuring stability while covering a distance of 2 meters.
\end{enumerate}

\begin{table}[!htb]
    \centering
    \begin{tabular}{rccc}
        \toprule
        Task & Distance (m) & Success (\%) & Tracking Error (m/s) \\ 
        \midrule
        Walking & 2.0 m  & 100 & 76.4 \\
        Avoidance & 1.85 m & 80 &  96.2 \\
        Climbing & 1.37 m & 95 & 70.6 \\
        \bottomrule
    \end{tabular}
    \caption{The results obtained from 20 simulated runs, with each row entry indicating the following metrics: average distance traversed, success rate percentage, and tracking error, respectively.}
    \label{table:experiment}
\end{table}

\subsection{Quantitative Evaluation in Navigation Experiment}

Reviewing Table \ref{table:experiment}, QTOS demonstrates a success rate of $91.7\%$ across the three navigation tasks. However, there is a noticeable performance gap between the Avoidance task and the Walking task, indicating a challenge in controller tracking. In the Avoidance task, the trajectory stitching within the global planner remains unaffected by sudden changes in direction, unlike the body controller. Consequently, the global planner tends to advance ahead of the current controller plan, resulting in larger-than-normal tracking errors. Nevertheless, the gait plans generated by QTOS exhibit sufficient stability to enable the robot's successful completion of the task. It is worth noting that in most of the simulated runs for the Avoidance task, failure occurred because the controller lacked the agility to reactively push the robot away from the map's boundary, or the gap between the wall and the edge of the map was too small for the robot to fit through. In the Climbing task, it was evident that the failures typically occurred when one of the robot's legs made unmodeled contact with the terrain, leading to its worst performance in terms of the average distance traversed. This highlights a significant flaw in TOWR's constraint formulation, which could potentially result in non-recoverable drift. However, since QTOS requires long-distance plans to be incrementally solved with stitched gait plans, the likelihood of terrain clipping along the trajectory plan is significantly reduced.

\subsection{Stitched Trajectory Tracking Experiment}

In this section, we evaluate the reliability of stitched trajectory gait planning over a 10-minute time period for a walking task. We track both the reference center of mass and the robot's center of mass positions to assess the degree of system drift and the performance of the controller. This experiment focuses on addressing two crucial questions: (1) Can the optimizer effectively handle continuously fed stitched trajectory plans? (2) Can the controller reasonably track trajectories over an extended period of time and distance without encountering significant drift?

\begin{figure}[!htb]
  \includegraphics[width=\linewidth]{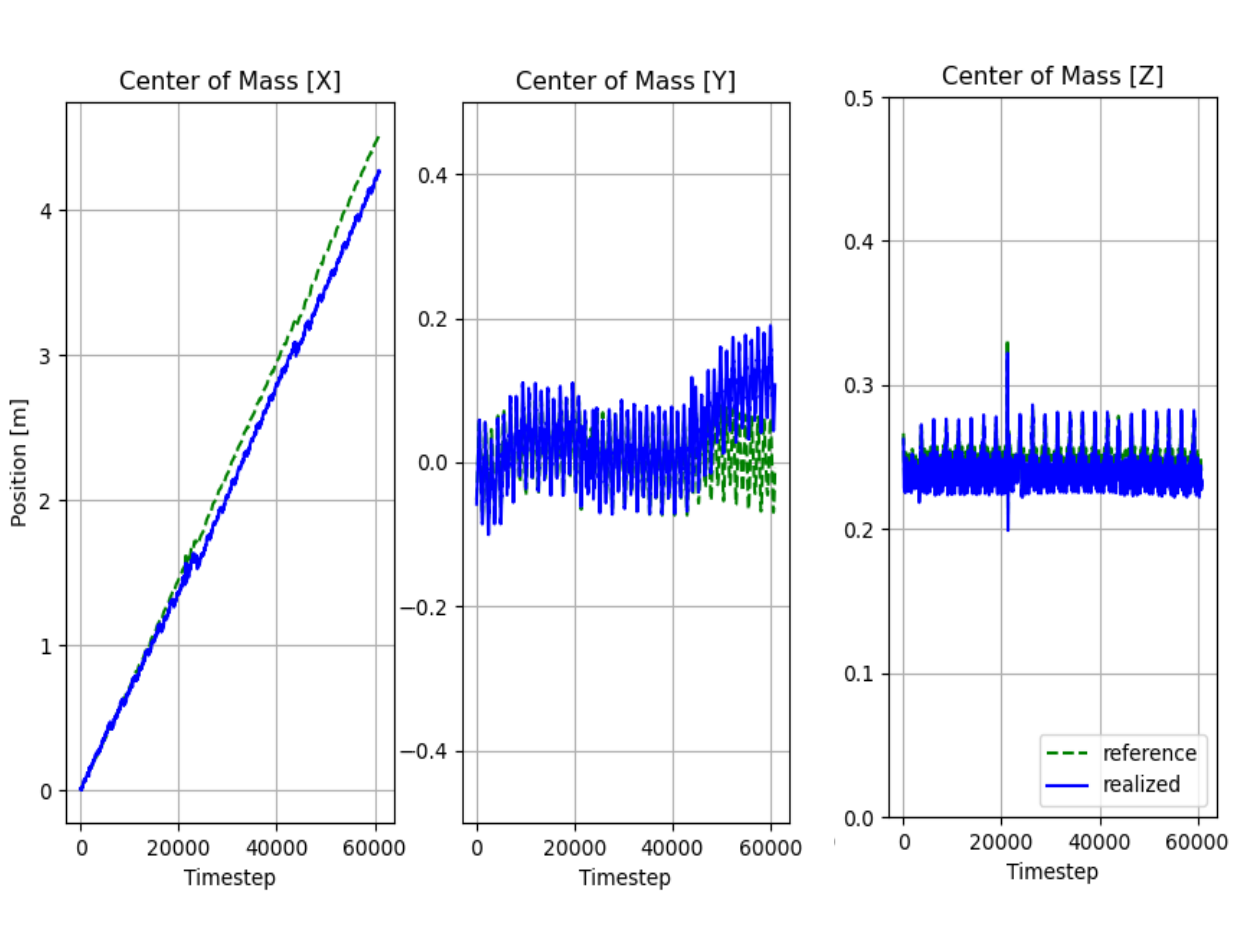}
  \caption{The reference and the robot's center of mass along the planned trajectory. The middle plot illustrates how the joint-velocity controller effectively corrects minor perturbations along the trajectory. The three plots were generated from QTOS.}
  \label{fig:tracking_graph}
\end{figure}
\subsection{Quantitative Evaluation of Stitched Trajectory Tracking}

We notice that the tracking error maintains a constant rate of change over the distance traveled by the robot. This suggests that the controller can handle a long-distance gait plan without significant loss in performance. Similarly, in Figure \ref{fig:tracking_graph}, the robot drifted to the left by approximately 0.1 meters over the 10-minute period, while the planned trajectory advanced about 0.2 meters ahead of the center of mass along the x-axis. Overall, the drift in the system can be compensated for by the controller, especially between time steps 0 to 20k, where the joint-velocity controller corrects the slight leftward drift. However, around time step 40k, the robot starts to exhibit significant leftward drift, indicating model alignment errors between the optimizer and the robot’s true model. From these observations, QTOS verifies that continuously fed stitched trajectory plans produced by TOWR are feasible and perform well in walking terrain.

\section{Conclusion}

In this paper, we introduced QTOS, an end-to-end software stack for quadruped locomotion. QTOS fills a crucial gap in the open-source quadruped community by integrating motion planners, solvers, control strategies, and a robot interface. This integration results in a user-friendly framework capable of generating stitched motion plans in simulation. The simulation experiments conducted in this study highlight QTOS's proficiency in performing walking, avoidance, and climbing tasks. Additionally, the stitched trajectory tracking experiments demonstrate QTOS's effectiveness in generating long-distance gait plans. In future iterations of QTOS, we plan to conduct extensive testing of generated gait plans with a physical SOLO12 robot. We also have intentions to enhance the framework's capabilities, either in the domain of reinforcement learning \cite{seo2023learning} \cite{tsounis2020deepgait}  or nonlinear optimization \cite{ahn2021versatile}, with the objective of producing more agile gait plans through improvements in the local planner module.
\newpage
\bibliographystyle{plain} 
\bibliography{QTOS} 

\end{document}